\documentclass[runningheads]{llncs}
\usepackage{graphicx}
\usepackage{setspace}
\usepackage{booktabs}
\usepackage{array}
\usepackage{threeparttable}
\usepackage{gensymb}
\usepackage{caption}
\usepackage{subfigure}
\usepackage{setspace}
\usepackage{cite}
\usepackage{makecell}
\usepackage{multirow}
\usepackage{tabularx}
\usepackage{color,soul}
\usepackage{amsmath}
\usepackage{amssymb}
\usepackage{color,soul}
\usepackage{algorithm}
\usepackage{algpseudocode}
\usepackage{hyperref}

\begin{document}
\title{
Conditional Diffusion Models for Weakly Supervised Medical Image Segmentation
}
\author{Xinrong Hu\inst{1}
\and Yu-Jen Chen\inst{2}
\and Tsung-Yi Ho\inst{3}
\and Yiyu Shi\inst{1}}

\authorrunning{X. Hu et al.}

\institute{University of Notre Dame, Notre Dame, IN, USA\\ 
\email{\{xhu7, yshi4\}@nd.edu}
\and National Tsing Hua University, Taiwan
\and
The Chinese University of Hong Kong, Hong Kong}
\titlerunning{Conditional Diffusion Models for WSSS}
%
%
\maketitle              
\begin{abstract}

Recent advances in denoising diffusion probabilistic models have shown great success in image synthesis tasks.
While there are already works exploring the potential of this powerful tool in image semantic segmentation, its application in weakly supervised semantic segmentation (WSSS) remains relatively under-explored.  
Observing that conditional diffusion models (CDM) is capable of generating images subject to specific distributions, in this work, we utilize category-aware semantic information underlied in CDM to get the prediction mask of the target object with only image-level annotations.
More specifically, we locate the desired class by approximating the derivative of the output of CDM w.r.t the input condition.
Our method is different from previous diffusion model methods with guidance from an external classifier, which accumulates noises in the background during the reconstruction process.
Our method outperforms state-of-the-art CAM and diffusion model methods on two public medical image segmentation datasets, which demonstrates that CDM is a promising tool in WSSS. 
Also, experiment shows our method is more time-efficient than existing diffusion model methods, making it practical for wider applications. 
The codes are available at \href{https://github.com/xhu248/cond_ddpm_wsss}{https://github.com/xhu248/cond\_ddpm\_wsss}

\end{abstract}

\section{Introduction}
\label{sec:introdcution}
Medical image segmentation is always a critical task as it can be used for disease diagnosis, treatment planning, and anomaly monitoring.
Weakly supervised semantic segmentation attracts significant attention from medical image community since it greatly reduces the cost of dense pixel-wise labeling to get segmentation mask. 
In WSSS, the training labels are usually easier and faster to obtain, like image-level tags, bounding boxes, scribbles, or point annotations.
This work only focuses on WSSS with image-level tags, like whether a tumor presents or not.
In this field, previous WSSS works\cite{izadyyazdanabadi2018weakly, patel2022weakly} are dominated by class activation map (CAM)\cite{zhou2016learning} and its variants\cite{selvaraju2017grad, chattopadhay2018grad, wang2020score, chen2023ame}, which was firstly introduced as a tool to visualize saliency maps when making a class prediction. 

\begin{figure}
    \centering
    \includegraphics[width=\linewidth]{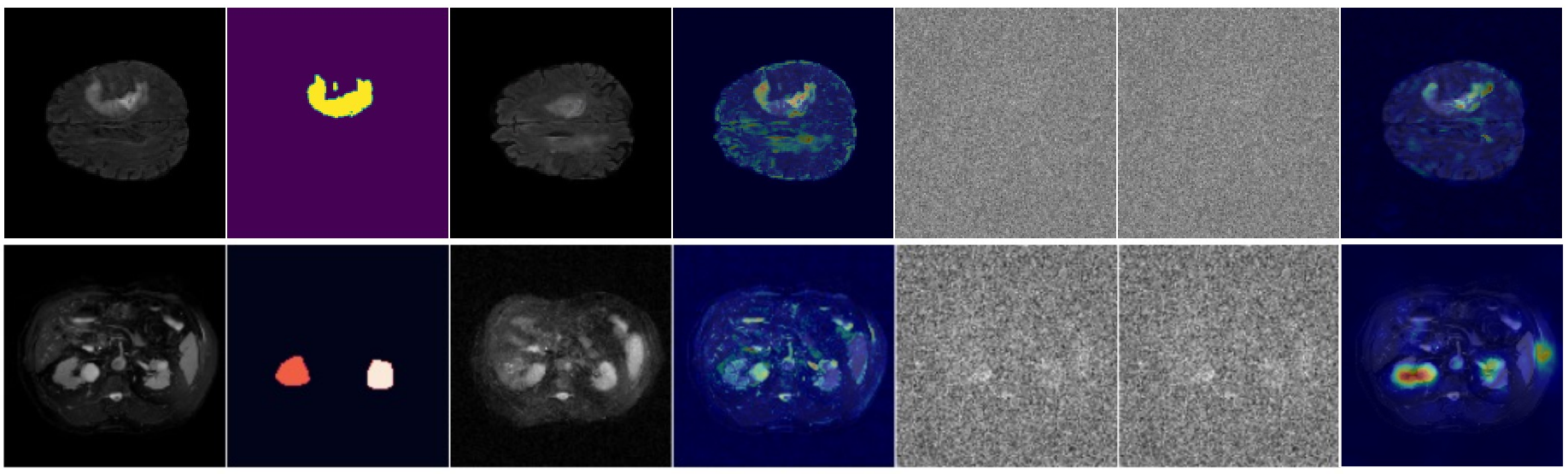}
    \caption{Intuition behind our CDM based method. From left to right, each column represents original images, ground truth, reconstructed images with negative guidance, saliency map generated by guided diffusion model, images with one step of denoising conditioned on positive label and negative label, and saliency map generated by our method. Images are from BraTS and CHAOS dataset.}
    \label{fig:mtv}
\end{figure}

Meanwhile, denoising diffusion models \cite{ho2020denoising, dhariwal2021diffusion} demonstrate superior performance in image synthesis than other generative models. 
Also, there are several works exploring the application of diffusion models to semantic segmentation in natural images \cite{baranchuklabel, graikos2022diffusion}and medical images\cite{wyatt2022anoddpm, pinaya2022fast, wolleb2022diffusion1, wolleb2022diffusion}. 
To the best of our knowledge, Wolleb et al.\cite{wolleb2022diffusion} is the only work that introduces diffusion models to pixel-wise anomaly detection with only classification labels. 
They achieve this by utilizing an external classifier trained with image-level annotations to guide the reverse Markov chain.
By passing the gradient of the classifier, the diffusion model gradually removes the anomaly areas during the denoising process and then obtains the anomaly map by comparing the reconstructed and original images
However, this approach is based on the hypothesis that the classifier can accurately locate the target objects and that the background is not changed when removing the noise. 
This assumption does not always hold, especially when the distribution of positive images is diverse, and the reconstruction error can also be accumulated after hundreds of steps. 
As shown in Fig. \ref{fig:mtv}, the reconstructed images guided by the gradient of non-kidney not only remove the kidney area but also change the content in the background.
Another limitation of this method is the long inference time required for a single image, as hundreds of iterations are needed to restore the image to its original noise level. In contrast, CAM approaches need only one inference to get the saliency maps.
Therefore, there is ample room for improvement in using diffusion models for WSSS task.

In this work, we propose a novel WSSS framework with conditional diffusion models (CDM) as we observe that the predicted noises on different condition show difference.
Instead of completely removing the noises from images, we calculate the derivative of the predicted noise after a few stages with respect to conditions so that the related objects are highlighted in the gradient map with less background misidentified.
As the output of diffusion model is not differentiable with respect to the discrete condition input, we adopt the finite difference method, i.e., perturbing the condition embedding by a small amplitude and logging the change of the output with DDIM\cite{songdenoising} generative process.
In addition, our method does not require the full reverse denoising process for the noised images and may only need one or a few iterations. 
Thus the inference time of our method is comparable to that of CAM-based approaches.
We evaluate our methods on two different tasks, brain tumor segmentation and kidney segmentation, and provide the quantitative results of both CAM based and diffusion model based methods as comparison.
Our approach achieves state-of-the-art performance on both datasets, demonstrating the effectiveness of the proposed framework.
We also conduct extensive ablation studies to analyze the impact of various components in our framework and provide reasoning for each design.


\section{Methods}
\label{sec:methods}
\subsection{Training Conditional Denoising Diffusion Models}
Suppose that we have a sample $x_{0}$ from distribution $D(x|y)$, and $y$ is the condition. The condition $y$ can be various, like different modality\cite{tashiro2021csdi}, inpainting\cite{batzolis2021conditional} and low resolution images\cite{saharia2022image}. In this work, $y\in\left\{ y_{0}, y_{1}\right\}$ indicates the binary classification label, like brain CT scans without tumor vs. with tumor.
We then gradually add Guassian noise to the original image sample with different level $t \in \left \{0, 1, ..., T \right \}$ as
\begin{equation}
    q(x_{t}|x_{t-1}, y) := \mathcal{N}(x_{t}|y; \sqrt{1-\beta_{t}}x_{t-1}|y, \beta_{t}\bold{I})
\end{equation}
With fixed variance $ \left\{\beta_{1}, \beta_{2},..., \beta_{t} \right \}$, $x_{t}$ can be explicitly expressed by $x_{0}$,
\begin{equation}
    q(x_{t}|x_{0}, y) := \mathcal{N}(x_{t}|y; \sqrt{\bar{\alpha}_{t}}x_{0}|y, (1- \bar{\alpha}_{t})\bold{I})
    \label{formula2}
\end{equation}
where $\alpha_{t} := 1 - \beta_{t}$, $\bar{\alpha}_{t} := \prod_{s=1}^{t}\alpha_{s}$.

Then a conditional U-Net\cite{ronneberger2015u} $\epsilon_{\theta}(x, t, y)$ is trained to approximate the reverse denoising process, 
\begin{equation}
    p_{\theta}(x_{t-1}|x_{t}, y) := \mathcal{N}(x_{t-1};\mu_{\theta}(x_{t}, t, y), \Sigma_{\theta}(x_t, t, y))
\end{equation}
The variance $\mu_{\sigma}$ can be learnable parameters or a fixed set of scalars, and both settings achieve comparable results in \cite{ho2020denoising}.  As for the mean, after reparameterization with $x_{t} = \sqrt{\bar{\alpha}_{t}}x_{0} + \sqrt{1 - \bar{\alpha}_{t}}\epsilon$ for $\epsilon \sim \mathcal{N}(\bold{0}, I)$, the loss function can be simplified as:
\begin{equation}
    L:=\mathbb{E}_{x_{0}, \epsilon}\left\| \epsilon - \epsilon_{\theta}(\sqrt{\bar{\alpha}_{t}}x_{0} + \sqrt{1 - \bar{\alpha}_{t}}\epsilon, t, y) \right\|
\end{equation}
As for how to infuse binary condition y in the U-Net, we follow the strategy in \cite{dhariwal2021diffusion}, using a embedding projection function $e = f(y), f\in\mathbb{R} \rightarrow \mathbb{R}^{n}$, with n being the embedding dimension. Then the condition embedding is added to feature maps in different blocks. 
After training the denoising model, Tashiro et al.\cite{batzolis2021conditional} proved that the network can yield the desired conditional distribution $D(x|y)$ given condition $y$.

\begin{algorithm}
\caption{Generation of WSSS prediction mask using differentiate conditional model with DDIM sampling}\label{alg:cap}
\begin{algorithmic}
\State Input: input image $x$ with label $y_{1}$, noise level $Q$, inference stage $R$, noise predictor $\epsilon_{\theta}$, $\tau$
\State Output: prediction mask of label $y_{1}$
\For{\textbf{all} t from 1 to Q}
    \State $x_{t}\gets \sqrt{1-\beta_{t}}x_{t-1} + \beta_{t}*\mathcal{N}(\bold{0}, \bold{I})$
    \EndFor
\State $a \gets $ zeros($x$.shape)
\State $x'_{t} \gets x_{t}$.copy() 
\For{\textbf{all} t from Q to Q-R}
    \State $\hat{\epsilon}_{1} \gets \epsilon_{\theta}(x_{t}, t, f(y_{1})),\; \hat{\epsilon}_{0} \gets \epsilon_{\theta}(x'_{t}, t, \tau f(y_{1}) + (1-\tau) f(y_{0}))$
    \State $x_{t-1} \gets \sqrt{\bar{\alpha}_{t-1}}(\frac{x_{t} -\sqrt{1 - \bar{\alpha}_{t}}\hat{\epsilon}_{1} }{\sqrt{\bar{\alpha}_{t}}}) + \sqrt{1 - \bar{\alpha}_{t-1}}\hat{\epsilon}_{1}$
    \State $x'_{t-1} \gets \sqrt{\bar{\alpha}_{t-1}}(\frac{x'_{t} -\sqrt{1 - \bar{\alpha}_{t}}\hat{\epsilon}_{0} }{\sqrt{\bar{\alpha}_{t}}}) + \sqrt{1 - \bar{\alpha}_{t-1}}\hat{\epsilon}_{0}$
    \State $a\gets a + \frac{1}{1 - \tau}(x_{t-1} - x'_{t-1})$
    \EndFor
\State \textbf{return} a
\end{algorithmic}
\end{algorithm}

\subsection{Gradient Map w.r.t Condition}
Inspired by the finding in \cite{baranchuklabel} that the denoising model extracts semantic information when performing reverse diffusion process, we aim to get segmentation mask from the sample generated by single or just several reverse Markov steps with DDIM\cite{songdenoising}. 
The reason for using DDIM is that one can generate a sample $x_{t-1}$ from $x_{t}$ deterministically if removing the random noise term via:
\begin{equation}
    \label{formula:sample}
    x_{t-1}(x_{t}, t, y) = \sqrt{\bar{\alpha}_{t-1}}(\frac{x_{t} -\sqrt{1 - \bar{\alpha}_{t}}\hat{\epsilon}_{\theta}(x_{t}, y) }{\sqrt{\bar{\alpha}_{t}}}) + \sqrt{1 - \bar{\alpha}_{t-1}}\hat{\epsilon}_{\theta}(x_{t}, y)
\end{equation}
When given the same images at noise level Q, but with different conditions,  the noises predicted by the network $\epsilon_{\theta}$ are supposed to reflect the localization of target objects, that is equivalently $\left \| x_{t-1}(x_{t}, t, y_{1}) - x_{t-1}(x_{t}, t, y_{0})\right \|$. This idea is quite similar to \cite{wolleb2022diffusion} if we keep sampling $x_{t-1}$ until $x_{0}$. 
However, it is not guaranteed that the condition $y_{0}$ does not change background areas besides the object needed to be localized.
Therefore, in order to minimize the error caused by the generation process, we propose to visualize the sensitivity of $x_{t-1}(x_{t}, t, y_{1})$ w.r.t condition $y_{1}$, that is $\frac{\partial x_{t-1}}{\partial y}$.
Since the embedding projection function $f(\cdot)$ is not differentiable, we approximate the gradient using the finite difference method:
\begin{equation}
    \frac{\partial x_{t-1}(x_{t}, t, y)}{\partial y}  |_{y=y1} =\lim_{\tau\rightarrow 1} \frac{x_{t-1}(x_{t}, t, f(y_{1})) - x_{t-1}(x_{t}, t, \tau f(y_{1}) + (1-\tau) f(y_{0}))}{1 - \tau}
\end{equation}
in which, $\tau$ is the moving weight from $f(y_{1})$ towards $f(y_{0})$. The weight $\tau$ can not be too close to 1, otherwise there is no noticeable gap between $x_{t-1}$ and $x'_{t-1}$, and we find $\tau=0.95$ gives the best performance. 
Algorithm \ref{alg:cap} shows the detailed workflow of obtaining the segmentation mask of samples with label $y_{1}$. 
Notice that we iterate the process (\ref{formula:sample}) for R steps, and the default $R$ in this work is set as 10, much smaller than $Q=400$. 
The purpose of $R$ is to amplify the change of $x_{t-1}$ since the condition does not change the predicted noise a lot in one step. 

In addition, we find that the guidance of additional classifier can further boost the WSSS task, by passing the gradient $\hat{\epsilon} \gets \epsilon_{\theta}(x_{t}) - s\sqrt{1-\bar{\alpha}_{t}}\nabla_{x_{t}}logp_{\phi}(y|x_{t})$, where $p_{\phi}$ is the classifier and $s$ is the gradient scale.
Then, in Algorithm 1, $\hat{\epsilon}_{1}$ and $\hat{\epsilon}_{0}$ have additional terms guided by gradient of $y_{1}$ and $y_{0}$, respectively.
The ablation studies of related hyperparameters can be seen in Section \ref{sec:results}.

\section{Experiments and Results}
\label{sec:experiment}
\textbf{Brain Tumor Segmentation} BraTS(Brain Tumor Segmentation challenge)\cite{bakas2017advancing} contains 2,000 cases of 3D brain scans, each of which includes four different MRI modalities as well as tumor segmentation ground truth. In this work, we only use the FLAIR channel and treat all types of tumor as one single class. We divide the official training set into 9:1 for training and validation purpose, and evaluate our method on the official validation set. For preprocessing, we slice 
each volume into 2D images, following the setting in \cite{dey2021asc}. Then the total number of slices in training set is 193,905, and we report metrics on the 5802 positive samples in the test set 

\textbf{Kidney Segmentation} This task is conducted on dataset from ISBI 2019 CHAOS Challenge\cite{CHAOS2021}, which contains 20 volumes of T2-SPIR MR abdominal scans. CHAOS provides pixel-wise annotation for several organs, but we focus on the kidney.
We split the 20 volumes into four folds for cross-validation, and then decompose 3D volumes to 2D slices in every fold. 
In the test stage, we remove slices with area of interest taking up less than $5\%$ of the total area in the slice, in order to avoid the influence of extreme cases on the average results.

Only classification labels are used during training the diffusion models, and segmentation masks are used for evaluation in the test stage. For both datasets, we repeat the evaluation protocols for four times and report the average metrics and their standard deviation on test set.

\subsection{Implementation Details}
As for model architecture, we use the same setting as in \cite{graikos2022diffusion}. 
The diffusion model is based on U-Net with encoder and decoder consisting of resnet blocks.
We implement two different versions of the proposed method, one without 
classifier guidance and one with it. To differentiate the two in the experiments,
we continue to call them former CDM, and the latter CG-CDM. 
The classifier used in CG-CDM is the same as the encoder of the diffusion model.
We stop training the diffusion model after 50,000 iterations or 7 days, and the classifier is trained for 20,000 iterations.
We choose AdamW as the optimizer with learning rate being 1e-5 and 5e-5 for diffusion model and classifier. 
The batch sizes for both datasets are 2.
The implementation of all methods in this work is based on PyTorch library, and all experiments are run on a single NVIDIA RTX 2080Ti. 

\newcolumntype{Y}{>{\centering\arraybackslash}X}
\begin{table}[t]
    \footnotesize
    \centering
    \caption{Comparisons with state-of-the-art WSSS methods on BraTS dataset. ``CAM'' refers to CAM-based methods, ``DM'' means methods based on diffusion models, and ``FSL'' is short for fully supervised learning.}
    \label{table1}
    \begin{tabularx}{\linewidth}{c|c|Y|Y|Y|c}
    \toprule
 \multicolumn{1}{c|}{Types} & Methods &Dice$\uparrow$ & mIoU$\uparrow$ & HD95$\downarrow$ & infer time \\
 \hline

\multirow{4}{*}{CAM} &GradCAM\cite{selvaraju2017grad}   &0.235$\pm$0.075 &0.149$\pm$0.051 &44.4$\pm$8.9   &3.79s   \\
 &GradCAM++\cite{chattopadhay2018grad}   &0.281$\pm$0.084  &0.187$\pm$0.059  &32.6$\pm$6.0   &3.59s  \\
 &ScoreCAM\cite{wang2020score}   &0.303$\pm$0.053  &0.202$\pm$0.039  &32.7$\pm$2.1    &27.0s \\
 &LayerCAM\cite{jiang2021layercam}   &0.276$\pm$0.082  &0.184$\pm$0.058  &30.4$\pm$6.6    &4.07s  \\
 \hline
\multirow{3}{*}{DM} &CG-Diff\cite{wolleb2022diffusion}  &0.456$\pm$0.043 &0.325$\pm$0.036 &43.4$\pm$3.0 &116s  \\
&CDM  &0.525$\pm$0.057 &0.407$\pm$0.051 &26.0$\pm5.2$ &1.61s \\
&CG-CDM  &\textbf{0.563$\pm$0.023} &\textbf{0.450$\pm$0.012} &\textbf{19.2$\pm$3.2} &2.78s  \\

\hline
FSL &N/A & 0.902$\pm$0.028 & 0.814$\pm$0.023 & 8.1$\pm$1.9  &0.31s  \\

\bottomrule
\end{tabularx}
\end{table}

\begin{table}[t]
    \footnotesize
    \centering
    \caption{Comparison with state-of-the-art WSSS methods on CHAOS dataset.}
    \label{table2}
    \begin{tabularx}{0.8\linewidth}{c|c|Y|Y|Y}
    \toprule
 \multicolumn{1}{c|}{Types} & Methods &Dice$\uparrow$ & mIoU$\uparrow$ & HD95$\downarrow$ \\
 \hline

\multirow{4}{*}{CAM} &GradCAM\cite{selvaraju2017grad}   &0.105$\pm$0.017  &0.059$\pm$0.010  &33.9$\pm$5.1   \\
 &GradCAM++\cite{chattopadhay2018grad} &0.147$\pm$0.016  &0.085$\pm$0.010  &28.5$\pm$4.5     \\
 &ScoreCAM\cite{wang2020score}   &0.135$\pm$0.024  &0.078$\pm$0.015  &32.1$\pm$6.7   \\
 &LayerCAM\cite{jiang2021layercam}   &0.194$\pm$0.022 &0.131$\pm$0.018  &29.7$\pm$8.1     \\
 \hline
\multirow{3}{*}{DM} &CG-Diff\cite{wolleb2022diffusion}  &0.235$\pm$0.025 &0.152$\pm$0.020 &27.1$\pm$3.2 \\
&CDM  &0.263$\pm$0.028 &0.167$\pm$0.042 & 26.6$\pm$2.4  \\
&CG-CDM&\textbf{0.311$\pm$0.018} &\textbf{0.186$\pm$0.014} &\textbf{23.3$\pm$3.4} \\

\hline
FSL &N/A & 0.847$\pm$0.011 & 0.765$\pm$0.023 & 3.6$\pm$1.7  \\

\bottomrule
\end{tabularx}
    
\end{table}

\begin{figure}
    \centering
    \includegraphics[width=0.9\linewidth]{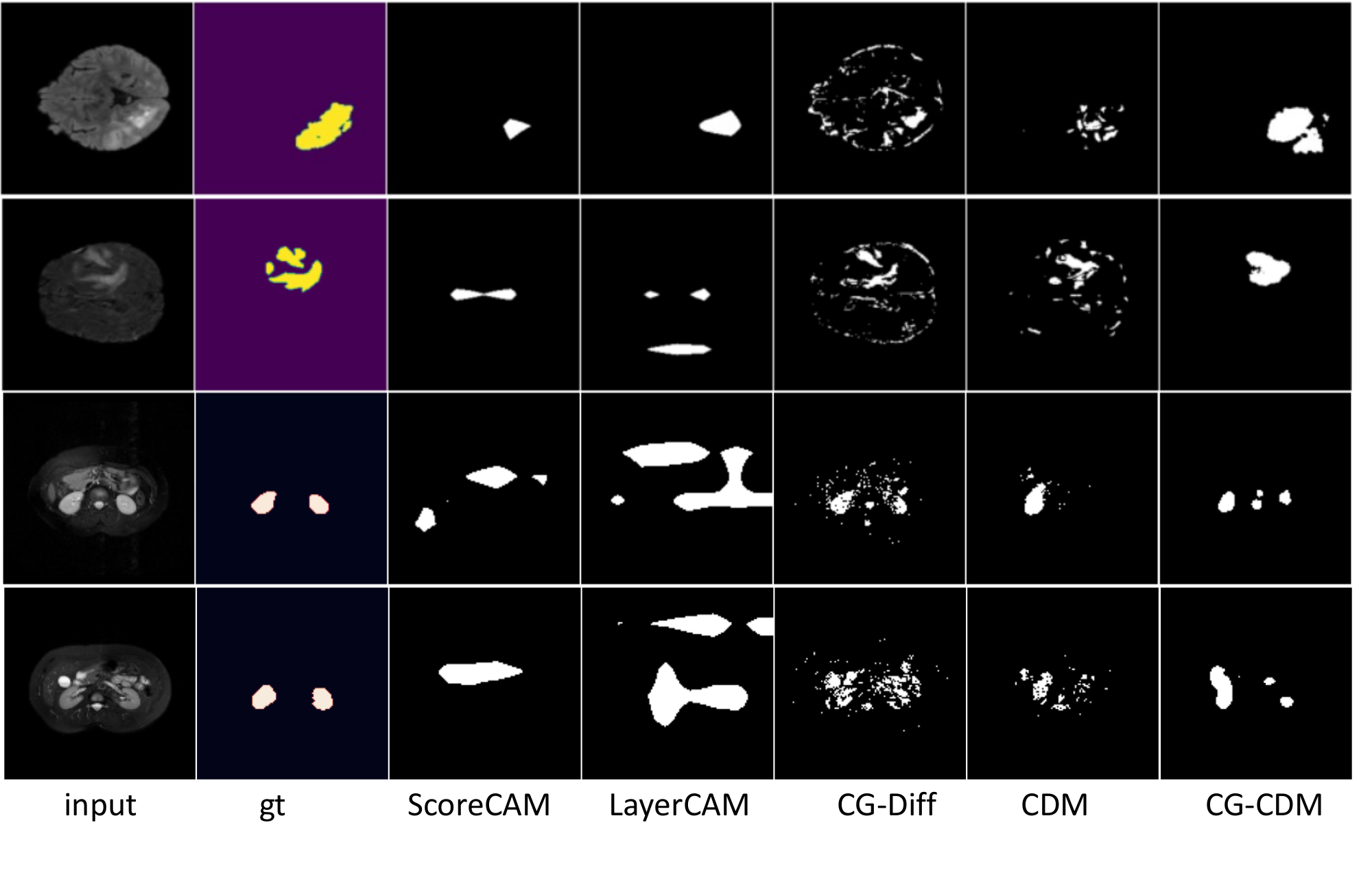}
    \caption{Visualization of WSSS segmentation masks using different methods on both BraTS and CHAOS dataset. The threshold is decided by average Otsu thresholding\cite{otsu1979threshold}.}
    \label{fig:my_label}
\end{figure}

\subsection{Results}
\label{sec:results}
\textbf{Comparison with state of the arts} We benchmark our methods against previous WSSS works on two datasets in Table\ref{table1} \& \ref{table2}, in terms of dice score, mean intersection over union (mIoU), and Hausdorff distance (HD95). For CAM based methods, we include the classical GradCAM\cite{selvaraju2017grad} and GradCAM++\cite{chattopadhay2018grad}, as well as two more recent methods, ScoreCAM\cite{wang2020score} and LayerCAM\cite{jiang2021layercam}. 
The implementation of these CAM approaches is based on the repository\cite{jacobgilpytorchcam}. 
For diffusion based methods, we include the only diffusion model for medical image segmentation in the WSSS literature, namely CG-Diff\cite{wolleb2022diffusion}.
We follow the default setting in \cite{wolleb2022diffusion}, setting noise level $Q=400$ and gradient scale $s=100$.
We also present the results under the fully supervised learning setting, which is the upper bond of all WSSS methods. 

From the results, we can make several key observations. 
Firstly, our proposed method, even without classifier guidance, outperform all other WSSS methods including the classifier guided diffusion model CG-Diff on both datasets for all three metrics. When classifier guidance is provided, the improvement gets even bigger, and CG-CDM can beat other methods regarding segmentation accuracy. 
Secondly, all WSSS methods have performance drop on kidney dataset compared with BraTS dataset.
This demonstrates that the kidney segmentation task is a more challenging task for WSSS than brain tumor task, which may be caused by the small training size and diverse appearance across slices in the CHAOS dataset.

\textbf{Time efficiency} Regarding inference time for different methods, as shown in Table\ref{table1}, both CDM and CG-CDM are much faster than CG-Diff.
The default noise level $Q$ is set as 400 for all diffusion model approaches, and our methods run 10 iterations during the denoising steps.
For all CAM-based approaches, we add augmentation smooth and eigen smooth suggested in \cite{jacobgilpytorchcam} to reduce noise in the prediction mask. 
This post-processing  greatly increases the inference time. 
Without the two smooth methods, the inference time for GradCAM is 0.031s, but the segmentation accuracy is significantly degraded.
Therefore, considering both inference time and performance, our method is a better option than CAM for WSSS.

\textbf{Ablation Studies} There are several important hyperparameters in our framework, noise level Q, number of iterations $R$, moving weight $\tau$, and gradient scale $s$. 
The default setting is CG-CDM on  BraTS dataset with $Q=400$, $R=10$, $\tau=0.95$, and $s=10$.
We evaluate the influence of one hyperparameter at a time by keeping other parameters at their default values. 
\begin{figure}
    \centering
    \includegraphics[width=1\linewidth]{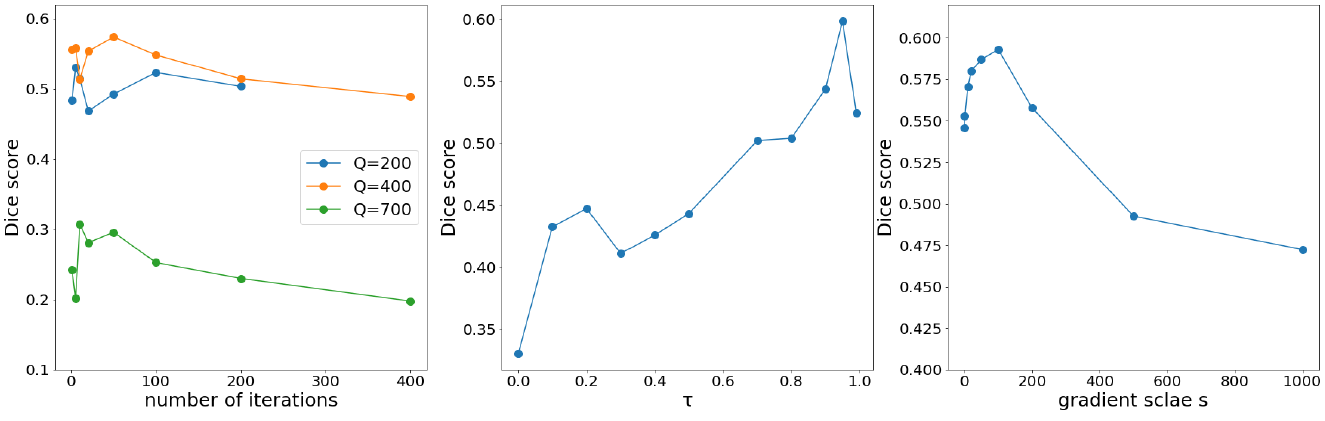}
    \caption{Ablation studies of hyper-parameters on BraTS dataset, including noise level $Q$, number of iterations $R$, moving weight $\tau$, and gradient scale $s$.}
    \label{fig:ablation}
\end{figure}
As illustrated in Fig. \ref{fig:ablation}, a few observations can be made: 
(1) Either too large or too small noise level can negatively influence the performance. 
When Q is small, most spatial information is still kept in $x_{t}$ and the predicted noise by diffusion model contains no semantic knowledge. 
When Q is large, most of the spatial information is lost and the predicted noise can be distracted from original structure.
Meanwhile, larger number of iterations can lightly improve the dice score at the beginning.
When $R$ gets too high, the error in the background is also accumulated after too many iterations.
(2) We try different $\tau$ in the range (0, 1.0).
Small $\tau$ leads to more noises in the background when calculating the difference in different conditions. 
On the other hand, as $\tau$ gets close to 1, the difference between $x_{t-1}$ and $x'_{t-1}$ becomes minor, and the gradient map mainly comes from the guidance of the classifier, making localization not so accurate. 
Thus, $\tau=0.95$ becomes the optimal choice for this task.
(3) As for gradient scale,  Fig. \ref{fig:ablation} shows that before $s=100$, larger gradient scale can boost the CDM, because at this time, the gradient from the classifier is at the same magnitude as the difference caused by the changed condition embedding.
When the guidance of the classifier becomes dominant, the dice score gets lower as the background is distorted by too large gradients.

\section{Conclusion}
In this paper, we present a novel weakly supervised semantic segmentation framework based on conditional diffusion models. 
Fundamentally, the essence of generative approaches on WSSS is maximizing the change in class-related areas while minimizing the noise in the background. Our methods are 
designed around this rule to enhance the state-of-the-art. First, 
existing work that utilizes a trained classifier to remove target objects leads to unpredictable distortion in other areas, thus we decide to iterate the reverse denoising process for as few steps as possible. Second, 
to amplify the difference caused by different conditions, we extract the semantic information from gradient of the noise predicted by the diffusion model.
Finally, this rule also applies to all other designs and choice of hyper-parameters in our framework.
When compared with latest WSSS methods on two public medical image segmentation
 datasets, our method shows superior performance regarding both segmentation accuracy and inference efficiency. 

\bibliographystyle{splncs04}
\bibliography{refs.bib}

\begin{thebibliography}{10}
\providecommand{\url}[1]{\texttt{#1}}
\providecommand{\urlprefix}{URL }
\providecommand{\doi}[1]{https://doi.org/#1}

\bibitem{bakas2017advancing}
Bakas, S., Akbari, H., Sotiras, A., Bilello, M., Rozycki, M., Kirby, J.S.,
  Freymann, J.B., Farahani, K., Davatzikos, C.: Advancing the cancer genome
  atlas glioma mri collections with expert segmentation labels and radiomic
  features. Scientific data  \textbf{4}(1),  1--13 (2017)

\bibitem{baranchuklabel}
Baranchuk, D., Voynov, A., Rubachev, I., Khrulkov, V., Babenko, A.:
  Label-efficient semantic segmentation with diffusion models. In:
  International Conference on Learning Representations

\bibitem{batzolis2021conditional}
Batzolis, G., Stanczuk, J., Sch{\"o}nlieb, C.B., Etmann, C.: Conditional image
  generation with score-based diffusion models. arXiv preprint arXiv:2111.13606
   (2021)

\bibitem{chattopadhay2018grad}
Chattopadhay, A., Sarkar, A., Howlader, P., Balasubramanian, V.N.: Grad-cam++:
  Generalized gradient-based visual explanations for deep convolutional
  networks. In: 2018 IEEE winter conference on applications of computer vision
  (WACV). pp. 839--847. IEEE (2018)

\bibitem{chen2023ame}
Chen, Y.J., Hu, X., Shi, Y., Ho, T.Y.: Ame-cam: Attentive multiple-exit cam for
  weakly supervised segmentation on mri brain tumor. arXiv preprint
  arXiv:2306.14505  (2023)

\bibitem{dey2021asc}
Dey, R., Hong, Y.: Asc-net: Adversarial-based selective network for
  unsupervised anomaly segmentation. In: Medical Image Computing and Computer
  Assisted Intervention--MICCAI 2021: 24th International Conference,
  Strasbourg, France, September 27--October 1, 2021, Proceedings, Part V 24.
  pp. 236--247. Springer (2021)

\bibitem{dhariwal2021diffusion}
Dhariwal, P., Nichol, A.: Diffusion models beat gans on image synthesis.
  Advances in Neural Information Processing Systems  \textbf{34},  8780--8794
  (2021)

\bibitem{jacobgilpytorchcam}
Gildenblat, J., contributors: Pytorch library for cam methods.
  \url{https://github.com/jacobgil/pytorch-grad-cam} (2021)

\bibitem{graikos2022diffusion}
Graikos, A., Malkin, N., Jojic, N., Samaras, D.: Diffusion models as
  plug-and-play priors. arXiv preprint arXiv:2206.09012  (2022)

\bibitem{ho2020denoising}
Ho, J., Jain, A., Abbeel, P.: Denoising diffusion probabilistic models.
  Advances in Neural Information Processing Systems  \textbf{33},  6840--6851
  (2020)

\bibitem{izadyyazdanabadi2018weakly}
Izadyyazdanabadi, M., Belykh, E., Cavallo, C., Zhao, X., Gandhi, S., Moreira,
  L.B., Eschbacher, J., Nakaji, P., Preul, M.C., Yang, Y.: Weakly-supervised
  learning-based feature localization for confocal laser endomicroscopy glioma
  images. In: Medical Image Computing and Computer Assisted
  Intervention--MICCAI 2018: 21st International Conference, Granada, Spain,
  September 16-20, 2018, Proceedings, Part II 11. pp. 300--308. Springer (2018)

\bibitem{jiang2021layercam}
Jiang, P.T., Zhang, C.B., Hou, Q., Cheng, M.M., Wei, Y.: Layercam: Exploring
  hierarchical class activation maps for localization. IEEE Transactions on
  Image Processing  \textbf{30},  5875--5888 (2021)

\bibitem{CHAOS2021}
Kavur, A.E., Gezer, N.S., Barış, M., Aslan, S., Conze, P.H., Groza, V., Pham,
  D.D., Chatterjee, S., Ernst, P., Özkan, S., Baydar, B., Lachinov, D., Han,
  S., Pauli, J., Isensee, F., Perkonigg, M., Sathish, R., Rajan, R., Sheet, D.,
  Dovletov, G., Speck, O., Nürnberger, A., Maier-Hein, K.H., {Bozdağı Akar},
  G., Ünal, G., Dicle, O., Selver, M.A.: {CHAOS Challenge - combined (CT-MR)
  healthy abdominal organ segmentation}. Medical Image Analysis  \textbf{69},
  101950 (Apr 2021). \doi{https://doi.org/10.1016/j.media.2020.101950},
  \url{http://www.sciencedirect.com/science/article/pii/S1361841520303145}

\bibitem{otsu1979threshold}
Otsu, N.: A threshold selection method from gray-level histograms. IEEE
  transactions on systems, man, and cybernetics  \textbf{9}(1),  62--66 (1979)

\bibitem{patel2022weakly}
Patel, G., Dolz, J.: Weakly supervised segmentation with cross-modality
  equivariant constraints. Medical Image Analysis  \textbf{77},  102374 (2022)

\bibitem{pinaya2022fast}
Pinaya, W.H., Graham, M.S., Gray, R., Da~Costa, P.F., Tudosiu, P.D., Wright,
  P., Mah, Y.H., MacKinnon, A.D., Teo, J.T., Jager, R., et~al.: Fast
  unsupervised brain anomaly detection and segmentation with diffusion models.
  In: Medical Image Computing and Computer Assisted Intervention--MICCAI 2022:
  25th International Conference, Singapore, September 18--22, 2022,
  Proceedings, Part VIII. pp. 705--714. Springer (2022)

\bibitem{ronneberger2015u}
Ronneberger, O., Fischer, P., Brox, T.: U-net: Convolutional networks for
  biomedical image segmentation. In: Medical Image Computing and
  Computer-Assisted Intervention--MICCAI 2015: 18th International Conference,
  Munich, Germany, October 5-9, 2015, Proceedings, Part III 18. pp. 234--241.
  Springer (2015)

\bibitem{saharia2022image}
Saharia, C., Ho, J., Chan, W., Salimans, T., Fleet, D.J., Norouzi, M.: Image
  super-resolution via iterative refinement. IEEE Transactions on Pattern
  Analysis and Machine Intelligence  (2022)

\bibitem{selvaraju2017grad}
Selvaraju, R.R., Cogswell, M., Das, A., Vedantam, R., Parikh, D., Batra, D.:
  Grad-cam: Visual explanations from deep networks via gradient-based
  localization. In: Proceedings of the IEEE international conference on
  computer vision. pp. 618--626 (2017)

\bibitem{songdenoising}
Song, J., Meng, C., Ermon, S.: Denoising diffusion implicit models. In:
  International Conference on Learning Representations

\bibitem{tashiro2021csdi}
Tashiro, Y., Song, J., Song, Y., Ermon, S.: Csdi: Conditional score-based
  diffusion models for probabilistic time series imputation. Advances in Neural
  Information Processing Systems  \textbf{34},  24804--24816 (2021)

\bibitem{wang2020score}
Wang, H., Wang, Z., Du, M., Yang, F., Zhang, Z., Ding, S., Mardziel, P., Hu,
  X.: Score-cam: Score-weighted visual explanations for convolutional neural
  networks. In: Proceedings of the IEEE/CVF conference on computer vision and
  pattern recognition workshops. pp. 24--25 (2020)

\bibitem{wolleb2022diffusion}
Wolleb, J., Bieder, F., Sandk{\"u}hler, R., Cattin, P.C.: Diffusion models for
  medical anomaly detection. In: Medical Image Computing and Computer Assisted
  Intervention--MICCAI 2022: 25th International Conference, Singapore,
  September 18--22, 2022, Proceedings, Part VIII. pp. 35--45. Springer (2022)

\bibitem{wolleb2022diffusion1}
Wolleb, J., Sandk{\"u}hler, R., Bieder, F., Valmaggia, P., Cattin, P.C.:
  Diffusion models for implicit image segmentation ensembles. In: International
  Conference on Medical Imaging with Deep Learning. pp. 1336--1348. PMLR (2022)

\bibitem{wyatt2022anoddpm}
Wyatt, J., Leach, A., Schmon, S.M., Willcocks, C.G.: Anoddpm: Anomaly detection
  with denoising diffusion probabilistic models using simplex noise. In:
  Proceedings of the IEEE/CVF Conference on Computer Vision and Pattern
  Recognition. pp. 650--656 (2022)

\bibitem{zhou2016learning}
Zhou, B., Khosla, A., Lapedriza, A., Oliva, A., Torralba, A.: Learning deep
  features for discriminative localization. In: Proceedings of the IEEE
  conference on computer vision and pattern recognition. pp. 2921--2929 (2016)

\end{thebibliography}

\appendix
\section{Appendix}
%

%

\begin{figure}
    \centering
    \includegraphics[width=1\linewidth]{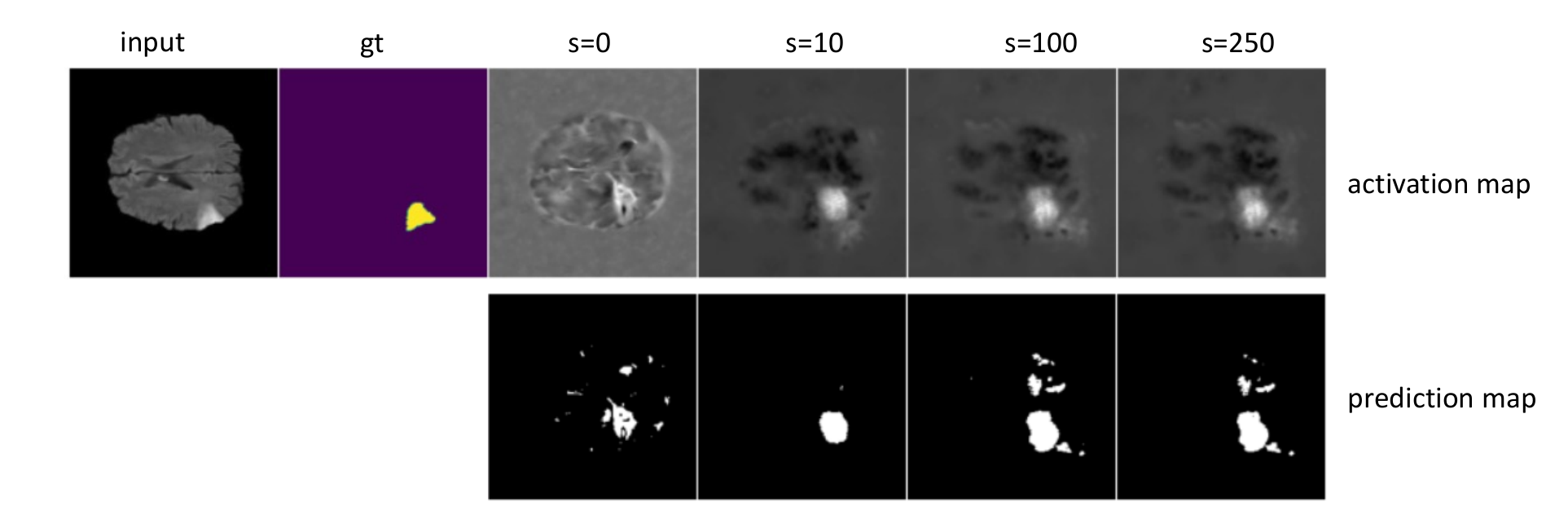}
    \caption{How gradient scale $s$ impacts the performance of our framework on BraTS dataset. When s=0, it means purely conditional diffusion model without guidance of external classifier. The activation map is the output $a$ from Algorithm 1. The prediction map is obtained by setting a average Otsu threshold.}
    \label{fig5}
\end{figure}

\begin{figure}
    \centering
    \includegraphics[width=\linewidth]{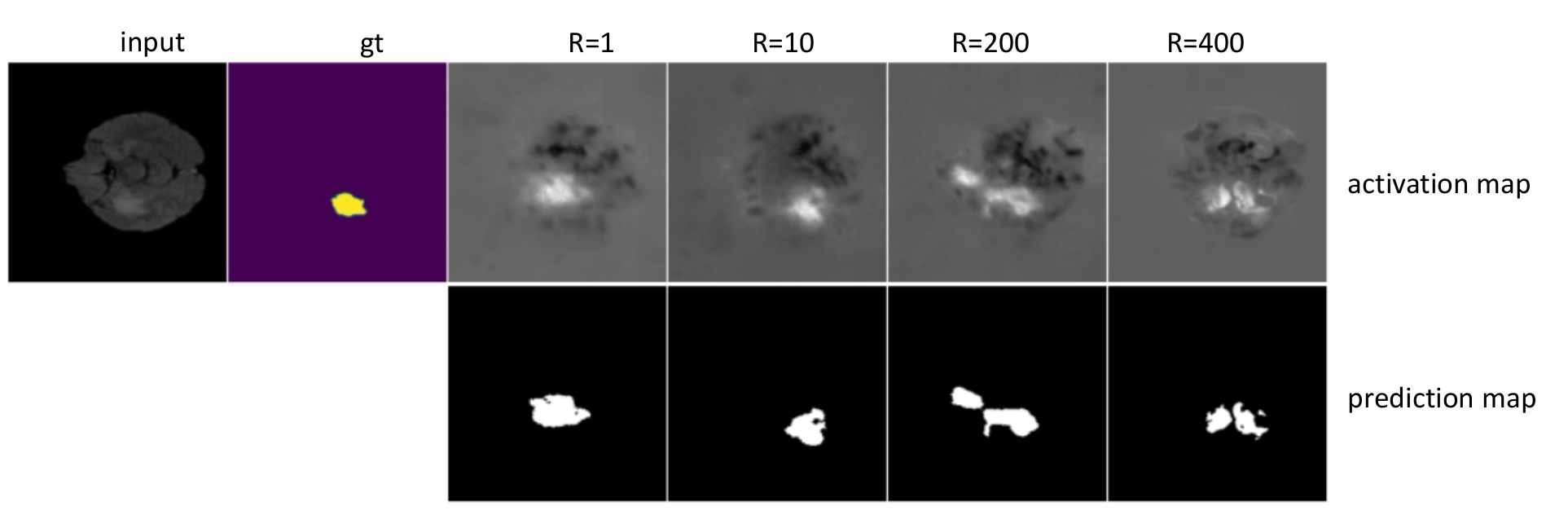}
    \caption{How inference iteration $R$ impacts the performance of our framework on BraTS dataset. The activation map is the output $a$ from Algorithm 1. The prediction map is obtained by setting a average Otsu threshold.}
    \label{fig5}
\end{figure}

\end{document}